\RequirePackage{fix-cm}
\documentclass[pdflatex,sn-basic]{sn-jnl}

\jyear{2026}

\usepackage[table]{xcolor}
\usepackage{graphicx}
\usepackage{multirow}
\usepackage{hhline}
\usepackage{amsmath}
\usepackage{booktabs}
\usepackage{tabularx}
\usepackage{array}
\usepackage{hyperref}
\usepackage[all]{hypcap}
\usepackage{cleveref}
\usepackage{placeins}
\usepackage{etoolbox}
\usepackage{microtype}

\newcommand{\dataname}{RAD}

\hypersetup{colorlinks=true,linkcolor=black,citecolor=black,urlcolor=blue!45!black}
\renewcommand{\Authorfont}{\normalfont\fontsize{11bp}{13bp}\selectfont\boldmath\titraggedcenter}
\renewcommand{\addressfont}{\normalfont\fontsize{9bp}{11bp}\selectfont\titraggedcenter}
\renewcommand{\abstractfont}{\normalfont\fontsize{9bp}{11bp}\selectfont\leftskip=0pt\rightskip=0pt}
\makeatletter
\patchcmd{\@maketitle}{\vskip21pt}{\vskip0pt}{}{}
\patchcmd{\@maketitle}{\vskip20pt}{\vskip12pt}{}{}
\patchcmd{\@maketitle}{\vskip20pt}{\vskip12pt}{}{}
\patchcmd{\@maketitle}{\vskip24pt}{\vskip10pt}{}{}
\patchcmd{\@maketitle}{\vskip24pt}{\vskip10pt}{}{}
\patchcmd{\@maketitle}{\vskip36pt}{\vskip16pt}{}{}
\makeatother
\apptocmd{\thebibliography}{\interlinepenalty=10000\clubpenalty=10000\widowpenalty=10000}{}{}
\begin{document}

\title[\dataname{}]{\dataname{}: A Realistic Multi-View Benchmark for Pose-Agnostic Anomaly Detection}

\author[2]{\fnm{Xinhai} \sur{Chang}}\equalcont{These authors contributed equally to this work.}
\author[1]{\fnm{Kaichen} \sur{Zhou}}\equalcont{These authors contributed equally to this work.}
\author[2]{\fnm{Taewhan} \sur{Kim}}\equalcont{These authors contributed equally to this work.}
\author[3]{\fnm{Jiadong} \sur{Zhang}}\equalcont{These authors contributed equally to this work.}
\author[4]{\fnm{Yang} \sur{Cao}}
\author[5]{\fnm{Chufei} \sur{Peng}}
\author[1]{\fnm{Fangneng} \sur{Zhan}}
\author[6]{\fnm{Hao} \sur{Zhao}}
\author[2]{\fnm{Hao} \sur{Dong}}
\author[7]{\fnm{Kai Ming} \sur{Ting}}
\author*[8]{\fnm{Ye} \sur{Zhu}}\email{ye.zhu@ieee.org}

\affil[1]{\orgname{Massachusetts Institute of Technology}, \orgaddress{\city{Cambridge}, \country{USA}}}
\affil[2]{\orgname{Peking University}, \orgaddress{\city{Beijing}, \country{China}}}
\affil[3]{\orgname{Carnegie Mellon University}, \orgaddress{\city{Pittsburgh}, \country{USA}}}
\affil[4]{\orgname{Great Bay University}, \orgaddress{\city{Guangdong}, \country{China}}}
\affil[5]{\orgname{Harvard University}, \orgaddress{\city{Cambridge}, \country{USA}}}
\affil[6]{\orgname{Tsinghua University}, \orgaddress{\city{Beijing}, \country{China}}}
\affil[7]{\orgname{Nanjing University}, \orgaddress{\city{Nanjing}, \country{China}}}
\affil[8]{\orgname{Deakin University}, \orgaddress{\city{Geelong}, \country{Australia}}}

\abstract{Anomaly detection is essential for robotic perception and industrial inspection, yet most benchmarks are collected under controlled conditions with fixed viewpoints and stable illumination. These settings do not reflect robotic deployment, where camera pose, lighting, and surface reflectance vary continuously. We present RAD (Realistic Anomaly Detection), a robot-captured multi-view benchmark for pose-agnostic anomaly detection. RAD contains 4,047 RGB images from 13 everyday object categories, captured from 68 viewpoints per object with a Franka robotic arm and an RGB-D camera under uncontrolled lighting. The dataset covers four realistic defect types: scratched, missing, stained, and squeezed, with pixel-level annotations for localization. We benchmark representative 2D feature-based methods, 3D reconstruction pipelines, and vision-language models under a realistic setting in which test poses are unknown. Results show a clear gap between performance on conventional benchmarks and performance on RAD. Strong 2D feature-embedding methods remain the most reliable at image-level detection, whereas 3D approaches are more competitive for pixel-level localization but remain vulnerable to reflective materials, geometric symmetry, and sparse viewpoint coverage. Vision-language models perform poorly in both classification and localization. RAD provides a challenging testbed for robust robotic inspection and highlights open problems in jointly modeling appearance, geometry, and viewpoint uncertainty. Our website and code are available at \url{https://chang-xinhai.github.io/rad-website/}.}


\maketitle

\section{Introduction}

Anomaly detection is a fundamental capability for robotic perception and industrial inspection, enabling robots to identify defects and irregularities during manufacturing, assembly, and maintenance~\cite{sinha2024real}. As vision-based inspection systems are increasingly deployed on robotic platforms, anomaly detection must operate under unconstrained viewpoints, varying illumination, and complex material properties, conditions that are rarely encountered in existing benchmarks.

Despite significant progress, most widely used anomaly detection datasets, such as MVTec AD~\cite{bergmann2019mvtec} and VisA~\cite{zou2022spot}, are collected under controlled laboratory setups with fixed camera viewpoints, uniform backgrounds, and stable lighting. These assumptions simplify the detection problem but fail to reflect real robotic inspection scenarios, where cameras mounted on robotic arms observe objects from continuously changing poses. Even for defect-free objects, modest viewpoint changes can drastically alter appearance due to shading, specular reflections, or partial occlusions, making pose-agnostic anomaly detection fundamentally more challenging.

\begin{figure}[!tbp]
    \centering
    \includegraphics[width=\linewidth]{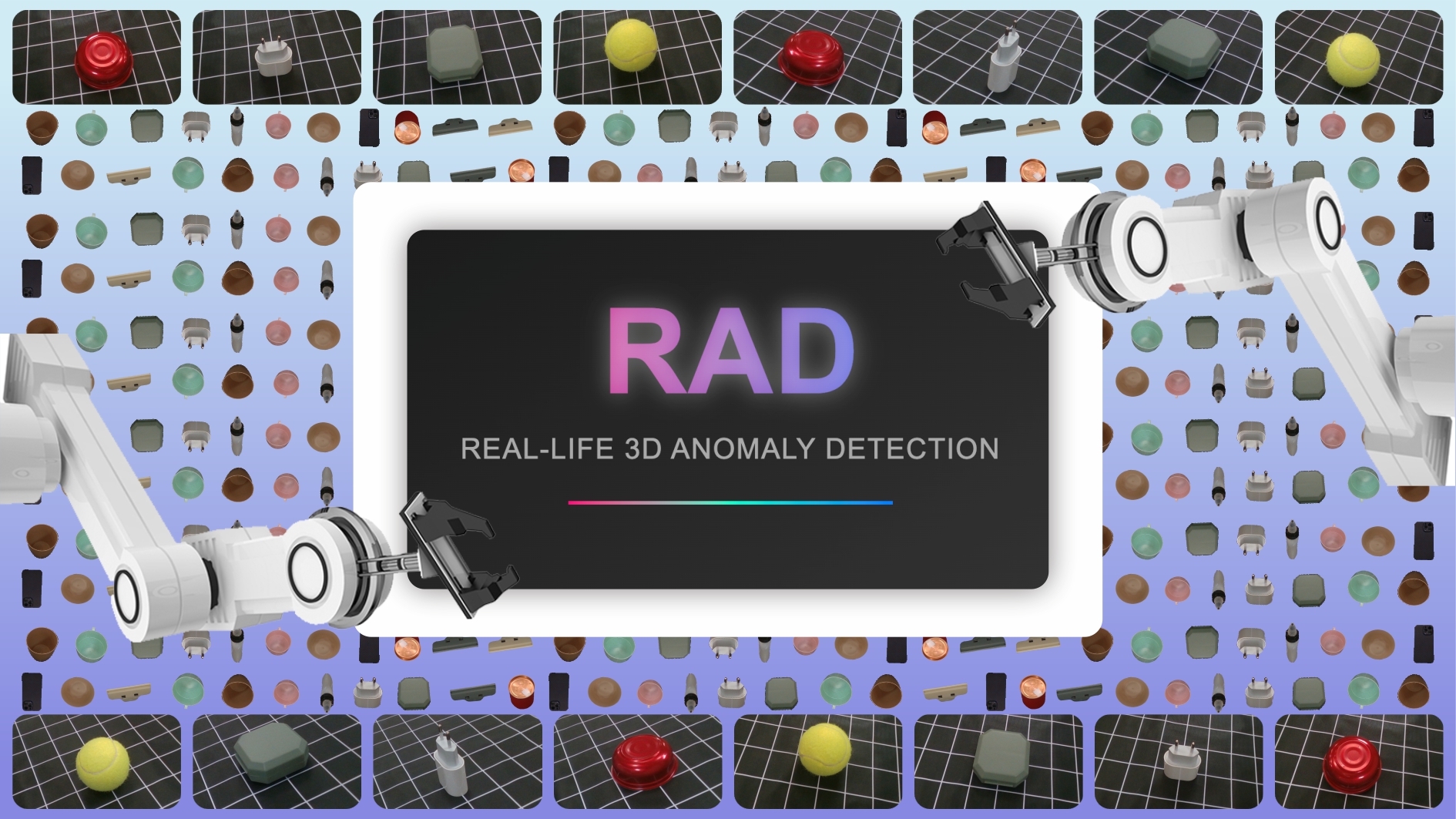}

    \caption{\textbf{Gallery of RAD.} RAD contains 13 industrial object categories captured from 68 robot viewpoints under uncontrolled illumination, introducing variations in pose, reflectance, and geometric symmetry that pose significant challenges for existing anomaly detectors.}
    \label{fig:teaser}

\end{figure}

Recent efforts have attempted to bridge this gap by incorporating multi-view data~\cite{visapp22,jezek2021deep}, 3D reconstruction~\cite{zhou2024pad,OmniPoseAD}, or vision-language models~\cite{bai2025qwen25vltechnicalreport,openai2024gpt4technicalreport}. While promising, these approaches introduce new assumptions that often break in practice. Geometry-based methods~\cite{park2021nerfies,zhou2026page,zhou2023dynpoint,zhou2026stream3d} rely on accurate pose estimation and novel-view synthesis, which are highly sensitive to reflective surfaces, geometric symmetry, and sparse view coverage. Zero-shot vision-language models, although powerful at high-level reasoning, lack pixel-level supervision and struggle to distinguish true defects from appearance changes caused by viewpoint variation. As a result, it remains unclear how existing methods perform under realistic robotic inspection conditions.

To address this gap, we introduce RAD (Realistic Anomaly Detection), a robot-captured, multi-view dataset and benchmark designed to reflect real-world inspection complexity. RAD contains 13 everyday object categories, each captured from 68 diverse robotic viewpoints under uncontrolled illumination, and includes four realistic defect types: scratched, missing, stained, and squeezed. The dataset emphasizes pose variation, reflective materials, and geometric symmetry, making anomaly detection substantially more challenging than in prior benchmarks.

Using RAD, we benchmark state-of-the-art 2D feature-based methods, 3D reconstruction pipelines, and vision-language models under a pose-agnostic setting. A key finding emerges: mature 2D feature-embedding methods consistently outperform recent 3D and vision-language approaches at the image level, despite the latter explicitly modeling geometry or semantics. While 3D methods achieve competitive pixel-level localization, their robustness is limited by pose ambiguity, reconstruction artifacts, and reflectance effects. Vision-language models perform poorly at both image- and pixel-level detection due to sensitivity to imaging conditions and lack of spatial supervision.

Our analysis identifies three fundamental challenges for realistic robotic inspection: reflective materials, geometric symmetry, and sparse viewpoint coverage. These results indicate that neither naive geometry augmentation nor zero-shot vision-language models are sufficient, motivating future methods that jointly reason over appearance and geometry with uncertainty awareness. Overall, RAD provides a challenging and realistic testbed for advancing pose-agnostic anomaly detection in robotics.

\section{Related Work}

\subsection{Anomaly Detection Datasets}

\begin{table}[!tbp]
\centering
\caption{\textbf{Comparison with industrial anomaly detection datasets.} RGB, D, PC, and NM denote color images, depth, point clouds, and normal maps. The upper panel reports data and sample counts; Views is the number of captured viewpoints. The lower panel reports capture and pose settings. Surface indicates transparent or specular objects.}
\label{tab:comparison}
\begingroup
\fontsize{8}{10}\selectfont
\setlength{\tabcolsep}{2pt}
\renewcommand{\arraystretch}{1.08}
\begin{tabular*}{\linewidth}{@{\extracolsep{\fill}}llcrrrr@{}}
\toprule
\textbf{Dataset} & \textbf{Data} & \textbf{Type} & \textbf{Classes} & \textbf{Normal} & \textbf{Anomalous} & \textbf{Views} \\
\midrule
\shortstack[l]{MVTec AD\\\cite{bergmann2019mvtec}} & RGB & real & 15 & 4,096 & 1,258 & 1 \\[1pt]
\shortstack[l]{MPDD\\\cite{jezek2021deep}} & RGB & real & 6 & 1,064 & 282 & 1 \\[1pt]
\shortstack[l]{VisA\\\cite{zou2022spot}} & RGB & real & 12 & 9,621 & 1,200 & 1 \\[1pt]
\shortstack[l]{MVTec 3D-AD\\\cite{visapp22}} & RGB/D/PC & real & 10 & 2,904 & 948 & 1 \\[1pt]
\shortstack[l]{Eyecandies\\\cite{bonfiglioli2022eyecandies}} & RGB/D/NM & synth & 10 & 13,250 & 2,250 & 1 \\[1pt]
\shortstack[l]{Real3D-AD\\\cite{liu2024real3d}} & RGB/D/PC & real & 12 & 652 & 602 & 360$^\circ$ \\[1pt]
\shortstack[l]{PAD\\\cite{zhou2024pad}} & RGB & synth & 20 & 4,960 & 4,412 & Around 20 \\[1pt]
\shortstack[l]{PAD\\\cite{zhou2024pad}} & RGB & real & 10 & 271 & 490 & Around 20 \\[1pt]
\shortstack[l]{Real-IAD\\\cite{wang2024real}} & RGB & real & 30 & 99,721 & 51,329 & 5 \\[1pt]
\shortstack[l]{SiM3D\\\cite{costanzino2025sim3d}} & RGB/D/PC & synth & 10 & 5,000 & 1,500 & 20 \\[1pt]
\midrule
\textbf{\dataname{} (ours)} & RGB & real & 13 & 1,224 & 2,823 & 68 \\
\bottomrule
\end{tabular*}
\par\vspace{8pt}
\begin{tabular*}{\linewidth}{@{\extracolsep{\fill}}llllcc@{}}
\toprule
\textbf{Dataset} & \textbf{View} & \textbf{Color} & \textbf{Surface} & \textbf{Train pose} & \textbf{Test pose} \\
\midrule
MVTec AD & Single & Diverse & Specular & 2D & 2D \\
MPDD & Single & Diverse & Specular & 2D & 2D \\
VisA & Single & Diverse & Specular & 2D & 2D \\
MVTec 3D-AD & Single & Diverse & Specular & 3D & 2D \\
Eyecandies & Single & Diverse & Transparency & 3D & 2D \\
Real3D-AD & Multi & Diverse & Transparency & 3D & 2D \\
PAD (synth) & Multi & Diverse & Specular & 3D & 2D \\
PAD (real) & Multi & Diverse & Specular & 3D & 2D \\
Real-IAD & Multi & Diverse & Specular & 3D & 3D \\
SiM3D & Multi & Diverse & Specular & 3D & 3D \\
\midrule
\textbf{\dataname{} (ours)} & Multi & Diverse & Specular & 3D & 3D \\
\bottomrule
\end{tabular*}
\endgroup
\end{table}

Anomaly detection is crucial for industrial automation and quality control~\cite{diers2023survey}. In 2D, benchmarks like MVTec AD~\cite{bergmann2019mvtec} established a standard by providing defect-free training images and diverse test anomalies. Extensions such as MVTec LOCO~\cite{bergmann2022beyond} and VisA~\cite{zou2022spot} added structural or logical anomalies and higher visual diversity. However, these single-view RGB datasets lack 3D geometry, hindering detection of spatially or viewpoint-dependent defects.

To address this limitation, several 3D anomaly detection datasets have been proposed. MVTec 3D-AD~\cite{visapp22} pairs RGB with aligned point clouds but uses a single fixed view under controlled lighting. MPDD~\cite{jezek2021deep} offers multi-view metal parts but at limited scale. Eyecandies~\cite{bonfiglioli2022eyecandies} uses synthetic candy-like objects and lacks realism. Real 3D-AD~\cite{liu2024real3d} operates in point clouds but requires impractical scanning setups. Real-IAD~\cite{wang2024real} captures 30 objects from only five fixed angles, offering sparse pose coverage. PAD~\cite{zhou2024pad} introduced pose-agnostic 3D anomaly detection using toy blocks, which limits industrial relevance. SiM3D~\cite{costanzino2025sim3d} bridges sim-to-real with multi-modal views but assumes perfect pose alignment.

As shown in \cref{tab:comparison}, \dataname{} is the first high-fidelity, multi-view, pose-agnostic benchmark targeting realistic robotic inspection of everyday objects. It uniquely emphasizes reflective surfaces, geometric symmetries, and viewpoint-dependent defect visibility. By capturing over 60 uncontrolled robotic viewpoints per object, \dataname{} offers a more authentic testbed for robust, geometry-aware, pose-invariant anomaly detection.

\begin{figure*}[!tbp]
    \centering
    \includegraphics[width=0.95\linewidth]{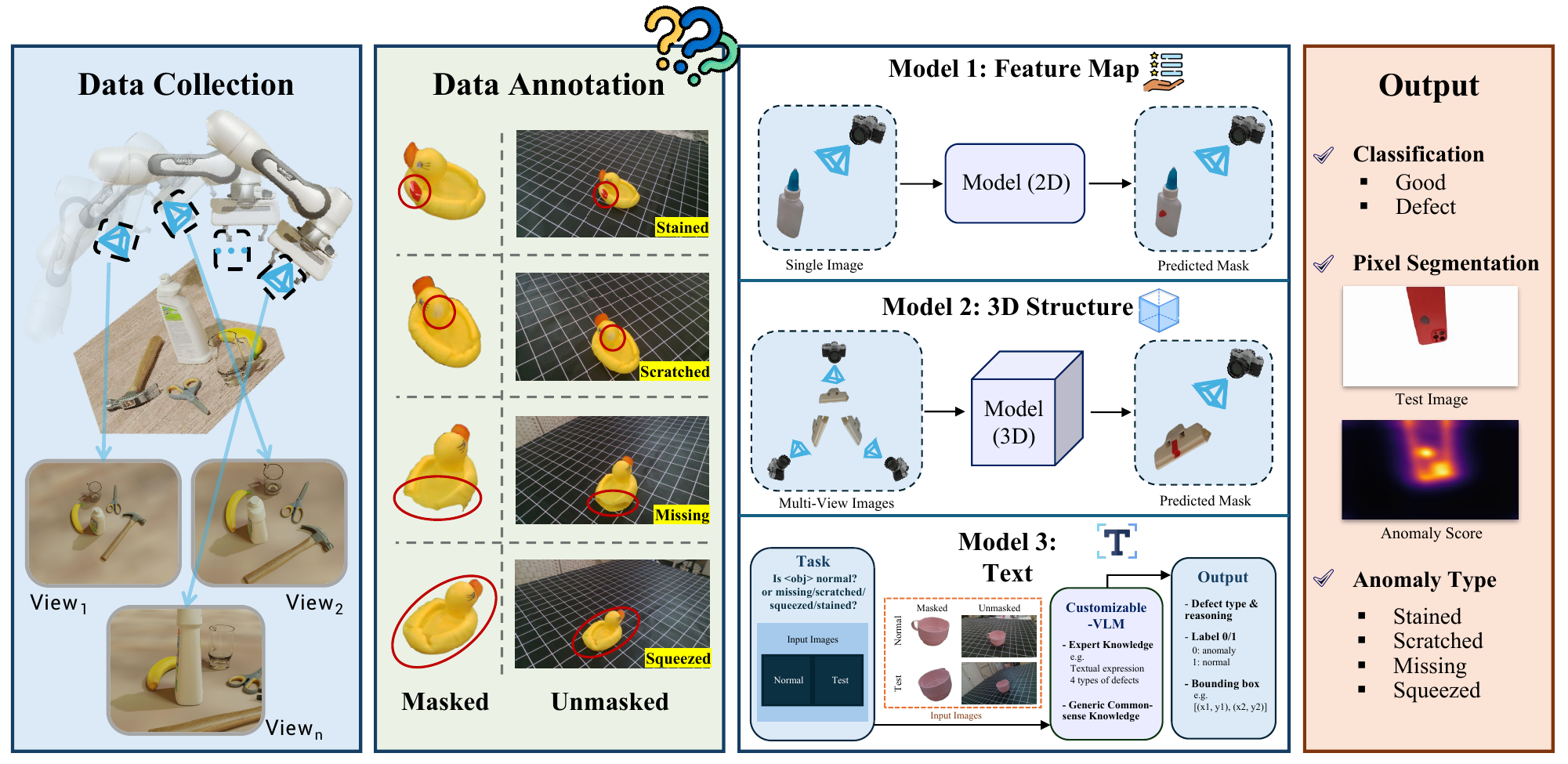}
    \caption{\textbf{Overview of the RAD anomaly detection benchmark pipeline.} The framework integrates robotic multi-view data collection, fine-grained data annotation, and three representative baseline families, including 2D, 3D, and vision-language pipelines, to support image-level classification, pixel-level localization, and defect-type analysis.}
    \label{fig:pipeline}
\end{figure*}

\subsection{Unsupervised Anomaly Detection}

Most unsupervised methods follow the MVTec AD setup: train on normal 2D RGB images and infer pixel-wise anomaly maps. They fall into two categories: reconstruction-based methods~\cite{zavrtanik2021draem,dehaene2020anomaly,liang2023omni,you2022unified,wyatt2022anoddpm,yan2021learning}, which detect reconstruction failures, and feature embedding-based methods~\cite{bergmann2020uninformed,salehi2021multiresolution,yi2020patch,massoli2021mocca,roth2022towards,gudovskiy2022cflow}, which flag deviations in learned feature spaces. To handle limited data, few-shot~\cite{jeong2023winclip,huang2022registration} and zero-shot approaches~\cite{zhou2023anomalyclip,AdaCLIP,qu2024vcp,ma2025aaclipenhancingzeroshotanomaly,gao2025adaptclipadaptingclipuniversal,fang2025afclipzeroshotanomalydetection} leverage pre-trained models, while anomaly synthesis~\cite{li2021cutpaste,yang2023memseg,jain2022synthetic} augments training with pseudo-defects.

With MVTec 3D-AD~\cite{visapp22}, multimodal methods emerged. AST~\cite{rudolph2023asymmetric} uses depth only for masking, remaining fundamentally 2D; M3AD~\cite{wang2023multimodal} fuses RGB and point cloud features explicitly. More recently, pose-agnostic detection gained traction. OmniPoseAD~\cite{OmniPoseAD} uses neural radiance fields~\cite{mildenhall2021nerf} and iNeRF~\cite{yen2021inerf} for pose refinement and view synthesis. SplatPose~\cite{SplatPose} replaces neural radiance fields with 3D Gaussian Splatting~\cite{kerbl20233d} for faster differentiable pose optimization. PIAD~\cite{yang2025piad} adds reflectance decomposition to 3D Gaussian Splatting, aiming for joint invariance to pose and illumination.

Concurrently, vision-language models enable zero-shot anomaly understanding. Commercial models like Qwen~\cite{bai2025qwen25vltechnicalreport} and GPT-4o~\cite{openai2024gpt4technicalreport} show promise in industrial contexts~\cite{Jiang2024MMAD}, while specialized methods~\cite{gu2023anomalygpt,xu2025towards,zhao2025omniaddetectunderstandindustrial,chao2025anomalyr1grpobasedendtoendmllm} use prompt engineering for competitive zero-shot performance. However, vision-language models typically output image-level classifications, lack pixel-level localization, and remain sensitive to clutter, texture details, and viewpoint changes.

\section{Dataset}

\subsection{Data Acquisition}

\textbf{Robotic capture setup.} RAD is collected using a Franka Emika Panda robotic arm equipped with an Intel RealSense D415 RGB-D camera (1280$\times$720). Each object is captured from 68 predefined viewpoints spanning a full 360$^\circ$ rotation. Objects are placed at the workspace center to ensure consistent imaging geometry. Camera poses are recorded during capture. The released RGB images are 1291$\times$721 pixels and are accompanied by anomaly masks and camera-pose metadata; raw depth observations are not released due to real-world depth noise.

\textbf{Advantages over conventional setups.} Compared to multi-camera rigs or rotating scanners used in prior datasets, such as PAD~\cite{zhou2024pad} and Real3D-AD~\cite{liu2024real3d}, the robotic arm enables dense multi-view coverage using a single sensor without object manipulation, closely reflecting practical industrial inspection scenarios while maintaining high repeatability.

\begin{figure}[!tbp]
    \centering
    \includegraphics[width=\linewidth]{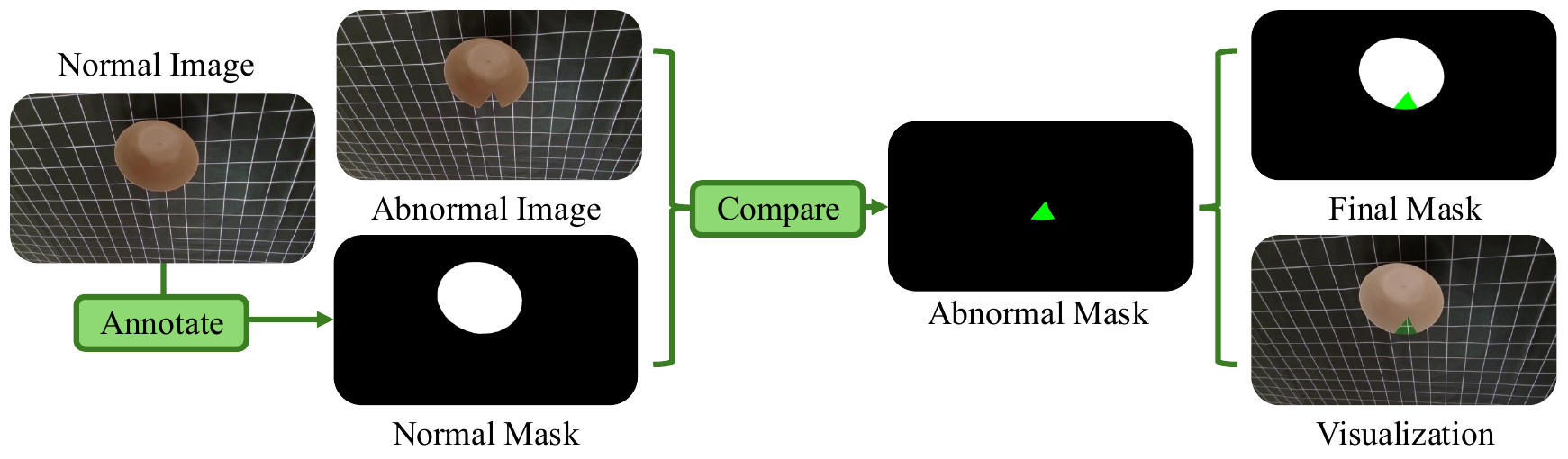}
    \caption{\textbf{Illustration of the annotation procedure.} Missing parts are annotated by comparing defective objects with their normal counterparts, producing precise binary masks for pixel-level evaluation.}
    \label{fig:data_annotation}

\end{figure}

\textbf{Annotation protocol.} Pixel-wise anomaly masks are generated by comparing defective images against corresponding normal views. Object regions are first annotated in a normal reference image, after which anomalies are manually labeled in the defective image using Adobe Photoshop, producing precise binary masks for scratches, stains, missing parts, and deformations (\cref{fig:data_annotation}).

\textbf{Anomaly types.} RAD includes four realistic defect types: (1) \textit{scratched}, surface incisions; (2) \textit{missing}, removed components; (3) \textit{stained}, localized discoloration; and (4) \textit{squeezed}, mechanical deformation. The released archives contain no squeezed samples for binderclip, box, charger, or tennisball; phonecase includes 68 squeezed images.

\subsection{Data Statistics}

\begin{figure}[!tbp]
    \centering
    \includegraphics[width=\linewidth]{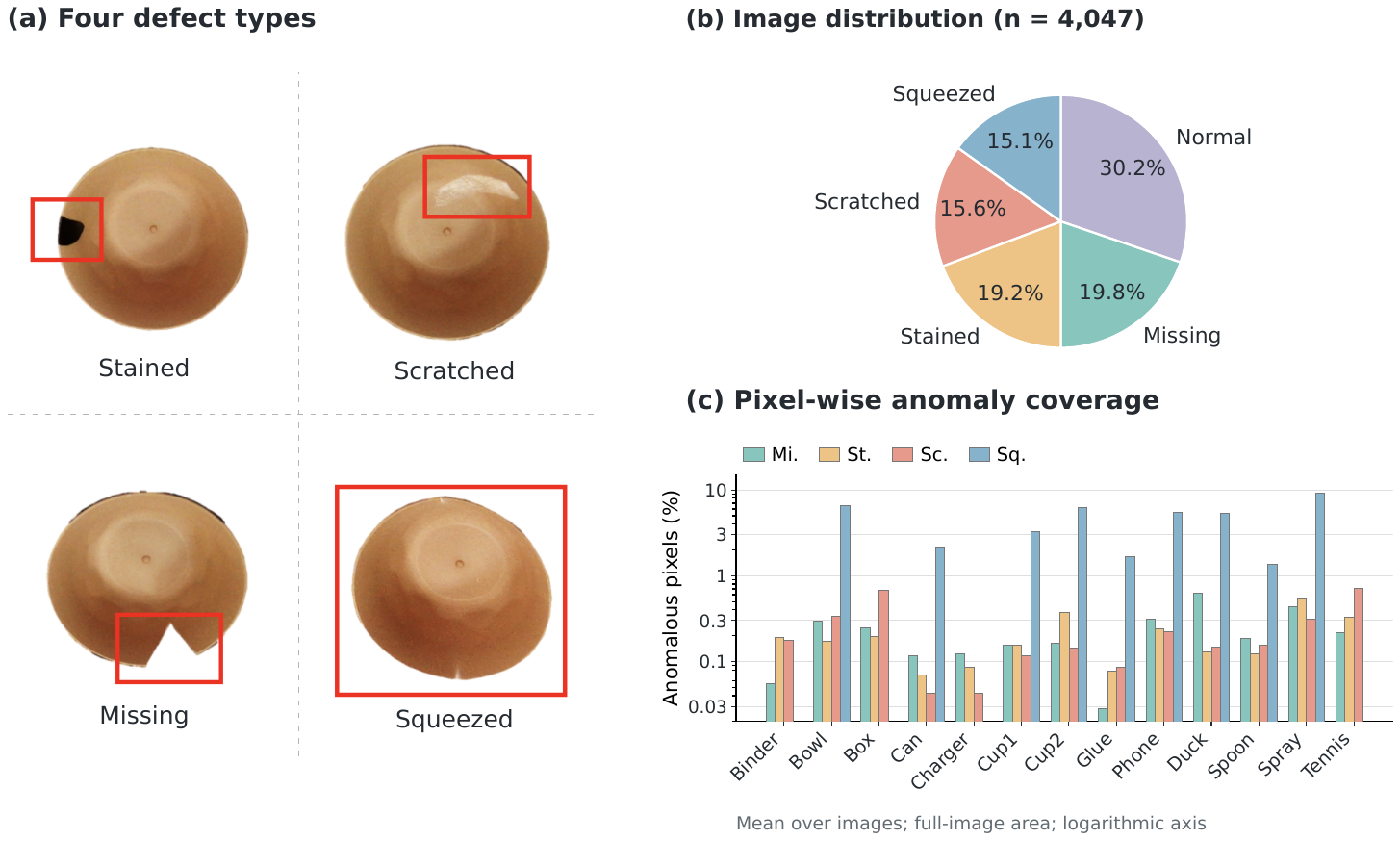}
    \caption{\textbf{Dataset metrics recomputed from the released archives.} (a) Examples of the four defect types. (b) Distribution of the 4,047 RGB images, including normal training and test views. (c) Mean anomalous-pixel coverage for each category and defect on a logarithmic axis: for each anomalous image, the number of positive ground-truth-mask pixels is divided by the full image area, then averaged over images from all corresponding instances. Mi., St., Sc., and Sq. denote missing, stained, scratched, and squeezed. Absent bars indicate no samples.}
    \label{fig:data_statistic}

\end{figure}

RAD consists of 4,047 RGB images across 13 semantic object categories represented by 18 physical-instance directories. The release contains 1,062 normal training images, 162 normal test images, and 2,823 anomalous test images. The anomalous images comprise 800 missing, 778 stained, 633 scratched, and 612 squeezed samples. Each anomalous image has a ground-truth mask; these 2,823 annotation files are not counted as additional RGB images. Each instance contributes 59 normal training views and 9 normal test views. As summarized in \cref{tab:data_statistic,fig:data_statistic}, the dataset exhibits diverse geometry, material reflectance, and defect scales. Both distributed archives have the same sample paths and include ground-truth masks and camera-pose metadata; the msk variant uses background-masked RGB images, while the nonmsk variant retains scene backgrounds.

\begin{table}[!tbp]
\setlength{\tabcolsep}{2pt}
\centering
\caption{\textbf{Dataset statistics.} ``Type'' denotes the number of distinct object instances per category. ``Attribute'' describes surface appearance: \textit{Single} = single-color non-specular, \textit{Multi.} = multi-color non-specular, \textit{Single/Specular} = single-color with specular reflections, and \textit{Multi./Specular} = multi-color with specular reflections. Defect columns: Miss. = missing, St. = stained, Sc. = scratched, and Sq. = squeezed. Normal includes both training and test views. Counts are RGB images; ground-truth masks are counted separately.}
\begin{tabular}{@{}>{\bfseries}l *{8}{c}@{}}
\toprule
\textbf{Category} & \textbf{Type} & \textbf{Attribute} & \textbf{Miss.} & \textbf{St.} & \textbf{Sc.} & \textbf{Sq.} & \textbf{Normal} & \textbf{Total} \\
\midrule
Binderclip & 2 & Single & 68 & 68 & 68 & 0 & 136 & 340 \\
Bowl & 1 & Single & 68 & 68 & 60 & 68 & 68 & 332 \\
Box & 1 & Single & 45 & 68 & 68 & 0 & 68 & 249 \\
Can & 1 & Multi./Specular & 41 & 68 & 4 & 68 & 68 & 249 \\
Rubberduck & 1 & Multi. & 68 & 60 & 58 & 68 & 68 & 322 \\
Spraybottle & 1 & Single/Specular & 52 & 62 & 58 & 68 & 68 & 308 \\
Cup1 & 1 & Single & 68 & 66 & 61 & 68 & 68 & 331 \\
Gluebottle & 2 & Multi./Specular & 53 & 21 & 24 & 68 & 136 & 302 \\
Charger & 1 & Multi./Specular & 66 & 66 & 5 & 0 & 68 & 205 \\
Cup2 & 3 & Single & 68 & 29 & 40 & 68 & 204 & 409 \\
Phonecase & 2 & Single & 68 & 68 & 68 & 68 & 136 & 408 \\
Spoon & 1 & Single & 68 & 66 & 51 & 68 & 68 & 321 \\
Tennisball & 1 & Multi. & 67 & 68 & 68 & 0 & 68 & 271 \\
\midrule
Total & 18 & -- & 800 & 778 & 633 & 612 & 1,224 & 4,047 \\
\bottomrule
\end{tabular}
\label{tab:data_statistic}

\end{table}

\section{\dataname{} Benchmark}

\subsection{Task: Pose-Agnostic Anomaly Detection}

In real robotic inspection, objects are observed from unknown and continuously varying viewpoints due to unconstrained robot motion. As a result, the same defect-free object can appear drastically different across views, making pose-agnostic anomaly detection essential.

The PAD framework~\cite{Kruse_2024_CVPR} defines this task as identifying and localizing defects when the object's pose is unknown and unconstrained. Unlike traditional anomaly detection, where training and test images are captured from fixed, aligned viewpoints, pose-agnostic anomaly detection requires the system to discern whether appearance differences stem from true anomalies or from pose changes. Formally, let the training set be
\[
R = \{(r_i, T_i)\}_{i=1}^N,
\]
where each normal image $r_i$ is paired with a known camera pose $T_i \in SE(3)$. At test time, a query image $q$ with unknown pose $T$ is provided.

The goal is to detect and localize anomalies without any anomalous training data. Pose-agnostic methods typically estimate an optimal pose $\hat{T}$ by aligning the query image to a 3D model $\mathcal{M}$ trained solely on normal views:
\begin{equation}
\hat{T} = \arg\min_{T \in SE(3)} \mathcal{L}_{\text{align}}\big(q, \mathcal{R}_{\mathcal{M}}(T)\big),
\end{equation}
where $\mathcal{R}_{\mathcal{M}}(T)$ renders the model from pose $T$ and $\mathcal{L}_{\text{align}}$ is a photometric or structural loss. The anomaly score is then computed in a deep feature space:
\begin{equation}
s(u) = \sum_\ell \big\| f^\ell(q)(u) - f^\ell(\mathcal{R}_{\mathcal{M}}(\hat{T}))(u) \big\|_2,
\end{equation}
with $f^\ell$ denoting the feature map at level $\ell$, $u$ a pixel index, and $s(u)$ the pixel-wise anomaly score.

\textbf{Why this is challenging.} Pose-agnostic anomaly detection is fundamentally harder than pose-aligned settings because models must distinguish true defects from appearance changes induced by viewpoint variation. This challenge is amplified by reflective materials, textureless surfaces, and geometric symmetries, which destabilize both pose estimation and appearance comparison. Sparse training views further exacerbate out-of-distribution effects at test time. These conditions reflect real-world robotic inspection and explain why methods performing well on pose-aligned benchmarks often degrade sharply in practice.

\subsection{Methods}

We benchmark representative anomaly detection approaches under pose-agnostic, real-world robotic inspection settings, covering three major paradigms: 2D feature-based methods, 3D reconstruction-based methods, and vision-language pipelines.

\textbf{2D feature-based.} We evaluate eight widely used unsupervised methods operating on 2D RGB images: CFlow~\cite{gudovskiy2022cflow}, EfficientAD~\cite{chan2022efficient}, FastFlow~\cite{yu2021fastflow}, PaDiM~\cite{defard2021padim}, PatchCore~\cite{roth2022towards}, Reverse Distillation~\cite{deng2022anomaly}, STFPM~\cite{wang2021student}, and UFlow~\cite{tailanian2024u}. In addition, we include three zero-shot CLIP variants: WinCLIP~\cite{jeong2023winclip}, AdaCLIP~\cite{AdaCLIP}, and VCPCLIP~\cite{qu2024vcp}, pretrained on VisA~\cite{visapp22}. All 2D baselines are implemented using standard Anomalib~\cite{akcay2022anomalib} configurations.

\textbf{3D reconstruction.} We benchmark SplatPose~\cite{kruse2024splatpose} and PIAD~\cite{yang2025piad}, which rely on recent 3D Gaussian Splatting techniques to build object-centric 3D representations from multi-view images. Given estimated camera poses, these methods render pose-aligned references and detect anomalies via geometry-consistent comparison. To better reflect real-world deployment, we use camera poses estimated by COLMAP~\cite{schoenberger2016sfm,schoenberger2016mvs} on unmasked images rather than ground-truth robotic poses.

\textbf{Vision-language pipelines.} Since off-the-shelf vision-language models do not support native pixel-level anomaly localization, we evaluate Qwen2.5-VL~\cite{bai2025qwen25vltechnicalreport} and ChatGPT-4o~\cite{openai2024gpt4technicalreport} using a three-step protocol: image-level anomaly classification, anomaly bounding-box prediction via prompting, and conversion of predicted boxes to binary masks for pixel-wise evaluation. The vision-language results serve as complementary baselines.

\subsection{Metrics}

Following prior work, we use the area under the receiver operating characteristic curve (AUROC) as the primary metric for both image-level anomaly classification and pixel-level segmentation, due to its robustness to class imbalance and threshold selection:
\begin{equation}
\mathrm{AUROC} = \int R_{\mathrm{TP}} \, dR_{\mathrm{FP}},
\end{equation}
where $R_{\mathrm{TP}}$ and $R_{\mathrm{FP}}$ denote the true positive rate and false positive rate, respectively.

For vision-language models, we additionally report \emph{type-wise AUROC}, which measures the model's ability to distinguish among different defect categories such as scratched, stained, missing, and squeezed.

\section{Experiments}

\subsection{Main Results}

We conduct a systematic evaluation of three representative families of anomaly detection methods on RAD: 2D feature-based approaches, 3D reconstruction-driven techniques, and vision-language models. Image-level and pixel-level ROC-AUC results are reported in \cref{tab:image_rocauc,tab:pixel_rocauc}.

\begin{table}[!tbp]
\centering
\caption{\textbf{Image-level AUROC on RAD.} The two panels cover all 13 semantic categories. Best results are in \textbf{bold}; runner-up results are \underline{underlined}.}
\label{tab:image_rocauc}
\begingroup
\fontsize{8.5}{10.5}\selectfont
\setlength{\tabcolsep}{2pt}
\renewcommand{\arraystretch}{1.06}
\begin{tabular*}{\linewidth}{@{\extracolsep{\fill}}llrrrrrrr@{}}
\toprule
\textbf{Method} & \textbf{Family} & \textbf{Binder} & \textbf{Bowl} & \textbf{Box} & \textbf{Can} & \textbf{Charger} & \textbf{Cup1} & \textbf{Cup2} \\
\midrule
CFlow & Feature & 0.737 & 0.468 & \textbf{0.973} & \textbf{0.938} & 0.493 & 0.470 & 0.653 \\
EfficientAD & Feature & \textbf{0.898} & 0.574 & 0.894 & 0.665 & \textbf{0.823} & \textbf{0.761} & \textbf{0.938} \\
FastFlow & Feature & 0.802 & \underline{0.815} & 0.871 & 0.758 & 0.681 & \underline{0.723} & 0.694 \\
PaDiM & Feature & 0.694 & 0.630 & 0.495 & 0.690 & 0.431 & 0.530 & 0.443 \\
PatchCore & Feature & \underline{0.832} & \textbf{0.896} & \underline{0.904} & \underline{0.884} & 0.601 & 0.639 & \underline{0.846} \\
Rev. Distill. & Feature & 0.806 & 0.756 & 0.867 & 0.700 & 0.702 & 0.709 & 0.692 \\
STFPM & Feature & 0.722 & 0.623 & 0.736 & 0.687 & 0.609 & 0.696 & 0.439 \\
UFlow & Feature & 0.686 & 0.684 & 0.729 & 0.621 & 0.389 & 0.604 & 0.649 \\
\addlinespace[2pt]
WinCLIP & Zero-shot & 0.648 & 0.790 & 0.872 & 0.658 & \underline{0.718} & 0.544 & 0.474 \\
AdaCLIP & Zero-shot & 0.620 & 0.602 & 0.647 & 0.557 & 0.662 & 0.694 & 0.591 \\
VCP-CLIP & Zero-shot & 0.574 & 0.585 & 0.896 & 0.532 & 0.681 & 0.693 & 0.637 \\
\addlinespace[2pt]
SplatPose & 3D & 0.701 & 0.528 & 0.493 & 0.521 & 0.461 & 0.494 & 0.532 \\
PIAD & 3D & 0.690 & 0.728 & 0.649 & 0.687 & 0.491 & 0.480 & 0.542 \\
\addlinespace[2pt]
Qwen2.5-VL & VLM & 0.282 & 0.632 & 0.000 & 0.240 & 0.000 & 0.711 & 0.222 \\
GPT-4o & VLM & 0.456 & 0.732 & 0.677 & 0.685 & 0.581 & 0.635 & 0.444 \\
\bottomrule
\end{tabular*}
\par\vspace{8pt}
\begin{tabular*}{\linewidth}{@{\extracolsep{\fill}}llrrrrrrr@{}}
\toprule
\textbf{Method} & \textbf{Family} & \textbf{Glue} & \textbf{Phone} & \textbf{Duck} & \textbf{Spoon} & \textbf{Spray} & \textbf{Tennis} & \textbf{Mean} \\
\midrule
CFlow & Feature & 0.633 & 0.507 & 0.486 & 0.458 & 0.784 & 0.437 & 0.618 \\
EfficientAD & Feature & \textbf{0.804} & \underline{0.929} & 0.717 & 0.806 & \textbf{0.882} & 0.963 & \underline{0.803} \\
FastFlow & Feature & 0.580 & 0.909 & 0.723 & \underline{0.895} & 0.747 & \underline{0.990} & 0.786 \\
PaDiM & Feature & 0.524 & 0.625 & 0.534 & 0.572 & 0.546 & 0.757 & 0.587 \\
PatchCore & Feature & \underline{0.722} & \textbf{0.986} & \textbf{0.947} & \textbf{0.959} & \underline{0.861} & \textbf{1.000} & \textbf{0.833} \\
Rev. Distill. & Feature & 0.691 & 0.911 & \underline{0.780} & 0.866 & 0.736 & 0.972 & 0.784 \\
STFPM & Feature & 0.572 & 0.706 & 0.742 & 0.679 & 0.526 & 0.914 & 0.679 \\
UFlow & Feature & 0.540 & 0.501 & 0.582 & 0.735 & 0.505 & 0.750 & 0.608 \\
\addlinespace[2pt]
WinCLIP & Zero-shot & 0.530 & 0.744 & 0.381 & 0.534 & 0.718 & 0.829 & 0.646 \\
AdaCLIP & Zero-shot & 0.485 & 0.715 & 0.555 & 0.545 & 0.803 & 0.818 & 0.627 \\
VCP-CLIP & Zero-shot & 0.714 & 0.625 & 0.658 & 0.749 & 0.513 & 0.860 & 0.663 \\
\addlinespace[2pt]
SplatPose & 3D & 0.562 & 0.436 & 0.474 & 0.456 & 0.599 & 0.532 & 0.524 \\
PIAD & 3D & 0.427 & 0.640 & 0.661 & 0.681 & 0.587 & 0.756 & 0.634 \\
\addlinespace[2pt]
Qwen2.5-VL & VLM & 0.230 & 0.265 & 0.240 & 0.450 & 0.598 & 0.670 & 0.305 \\
GPT-4o & VLM & 0.529 & 0.608 & 0.685 & 0.540 & 0.587 & 0.684 & 0.517 \\
\bottomrule
\end{tabular*}
\par\vspace{4pt}
\raggedright\fontsize{7.5}{9}\selectfont Binder: binder clip; Glue: glue bottle; Phone: phone case; Duck: rubber duck; Spray: spray bottle; Tennis: tennis ball.
\endgroup
\end{table}

\begin{table}[!tbp]
\centering
\caption{\textbf{Pixel-level AUROC on RAD.} The two panels cover all 13 semantic categories. Best results are in \textbf{bold}; runner-up results are \underline{underlined}.}
\label{tab:pixel_rocauc}
\begingroup
\fontsize{8.5}{10.5}\selectfont
\setlength{\tabcolsep}{2pt}
\renewcommand{\arraystretch}{1.06}
\begin{tabular*}{\linewidth}{@{\extracolsep{\fill}}llrrrrrrr@{}}
\toprule
\textbf{Method} & \textbf{Family} & \textbf{Binder} & \textbf{Bowl} & \textbf{Box} & \textbf{Can} & \textbf{Charger} & \textbf{Cup1} & \textbf{Cup2} \\
\midrule
CFlow & Feature & \underline{0.994} & 0.962 & 0.994 & \underline{0.989} & \textbf{0.998} & 0.972 & 0.977 \\
EfficientAD & Feature & \textbf{0.996} & \underline{0.966} & \underline{0.995} & \textbf{0.991} & \textbf{0.998} & 0.966 & 0.969 \\
FastFlow & Feature & 0.956 & 0.816 & 0.977 & 0.945 & 0.952 & 0.781 & 0.903 \\
PaDiM & Feature & 0.993 & 0.950 & 0.990 & 0.988 & 0.996 & 0.974 & 0.969 \\
PatchCore & Feature & 0.991 & 0.952 & 0.994 & 0.988 & 0.996 & 0.968 & 0.976 \\
Rev. Distill. & Feature & \underline{0.994} & 0.961 & 0.993 & 0.987 & \underline{0.997} & 0.980 & \underline{0.980} \\
STFPM & Feature & 0.990 & 0.952 & 0.984 & 0.980 & 0.991 & 0.960 & 0.957 \\
UFlow & Feature & 0.986 & 0.930 & 0.932 & 0.981 & 0.980 & 0.914 & 0.969 \\
\addlinespace[2pt]
WinCLIP & Zero-shot & 0.916 & 0.933 & 0.984 & 0.494 & 0.971 & 0.847 & 0.759 \\
AdaCLIP & Zero-shot & 0.984 & 0.902 & 0.979 & 0.984 & 0.977 & 0.932 & 0.940 \\
VCP-CLIP & Zero-shot & \underline{0.994} & \textbf{0.974} & \textbf{0.997} & \underline{0.989} & 0.996 & \textbf{0.984} & \textbf{0.984} \\
\addlinespace[2pt]
SplatPose & 3D & 0.990 & 0.965 & 0.984 & \textbf{0.991} & 0.993 & \underline{0.983} & 0.972 \\
PIAD & 3D & \underline{0.994} & 0.955 & 0.985 & 0.988 & 0.995 & 0.978 & 0.975 \\
\addlinespace[2pt]
Qwen2.5-VL & VLM & 0.525 & 0.573 & 0.636 & 0.507 & 0.514 & 0.505 & 0.517 \\
GPT-4o & VLM & 0.506 & 0.529 & 0.507 & 0.528 & 0.530 & 0.513 & 0.515 \\
\bottomrule
\end{tabular*}
\par\vspace{8pt}
\begin{tabular*}{\linewidth}{@{\extracolsep{\fill}}llrrrrrrr@{}}
\toprule
\textbf{Method} & \textbf{Family} & \textbf{Glue} & \textbf{Phone} & \textbf{Duck} & \textbf{Spoon} & \textbf{Spray} & \textbf{Tennis} & \textbf{Mean} \\
\midrule
CFlow & Feature & 0.989 & 0.981 & \textbf{0.988} & 0.985 & \underline{0.973} & 0.985 & \underline{0.984} \\
EfficientAD & Feature & 0.991 & 0.976 & 0.463 & 0.975 & 0.945 & 0.994 & 0.940 \\
FastFlow & Feature & 0.915 & 0.942 & 0.901 & 0.986 & 0.912 & 0.989 & 0.912 \\
PaDiM & Feature & 0.946 & \underline{0.988} & 0.980 & 0.992 & 0.974 & 0.992 & 0.972 \\
PatchCore & Feature & 0.929 & \textbf{0.991} & \underline{0.982} & \underline{0.994} & \textbf{0.985} & \textbf{0.996} & 0.978 \\
Rev. Distill. & Feature & 0.991 & 0.987 & 0.977 & \textbf{0.995} & 0.962 & 0.993 & \underline{0.984} \\
STFPM & Feature & 0.983 & 0.978 & 0.959 & 0.985 & 0.950 & 0.984 & 0.973 \\
UFlow & Feature & 0.986 & 0.942 & 0.955 & 0.967 & 0.930 & 0.960 & 0.951 \\
\addlinespace[2pt]
WinCLIP & Zero-shot & 0.878 & 0.616 & 0.957 & 0.848 & 0.872 & 0.987 & 0.853 \\
AdaCLIP & Zero-shot & 0.989 & 0.964 & 0.967 & 0.985 & 0.951 & 0.905 & 0.960 \\
VCP-CLIP & Zero-shot & \textbf{0.993} & 0.981 & 0.979 & 0.993 & 0.966 & \underline{0.995} & \textbf{0.987} \\
\addlinespace[2pt]
SplatPose & 3D & 0.990 & 0.976 & 0.971 & 0.992 & 0.937 & 0.988 & 0.977 \\
PIAD & 3D & 0.986 & 0.985 & 0.970 & 0.992 & 0.954 & 0.991 & \underline{0.984} \\
\addlinespace[2pt]
Qwen2.5-VL & VLM & 0.502 & 0.531 & 0.520 & 0.517 & 0.524 & 0.669 & 0.538 \\
GPT-4o & VLM & 0.516 & 0.506 & 0.528 & 0.511 & 0.552 & 0.510 & 0.517 \\
\bottomrule
\end{tabular*}
\par\vspace{4pt}
\raggedright\fontsize{7.5}{9}\selectfont Binder: binder clip; Glue: glue bottle; Phone: phone case; Duck: rubber duck; Spray: spray bottle; Tennis: tennis ball.
\endgroup
\end{table}

For image-level anomaly detection, 2D feature-based methods significantly outperform both 3D reconstruction and vision-language approaches. PatchCore achieves the highest average image-level AUROC of 0.833 and ranks first on seven object categories, including \textit{phonecase}, \textit{spoon}, and \textit{tennisball}. EfficientAD attains the second-best average score of 0.803. In contrast, 3D methods show notably weaker results, with average AUROC scores from 0.524 to 0.634. Vision-language models perform the worst, with average AUROC below 0.52 and frequent zero scores on categories such as \textit{can}, \textit{charger}, and \textit{spoon}.

For pixel-level segmentation, the performance gap narrows across method types. VCPCLIP achieves strong performance in categories such as \textit{box}, \textit{cup1}, and \textit{gluebottle}, leading all methods with an average pixel-level AUROC of 0.987. PatchCore, Reverse Distillation, and PIAD also maintain high segmentation accuracy, each exceeding 0.97 average AUROC. Notably, although 3D methods underperform at the image level, they remain competitive at the pixel level, with PIAD reaching 0.984 and SplatPose 0.977. Vision-language models remain ineffective at the pixel level, with average AUROC around 0.52.

\begin{figure*}[!tbp]
    \centering
    \includegraphics[width=\linewidth]{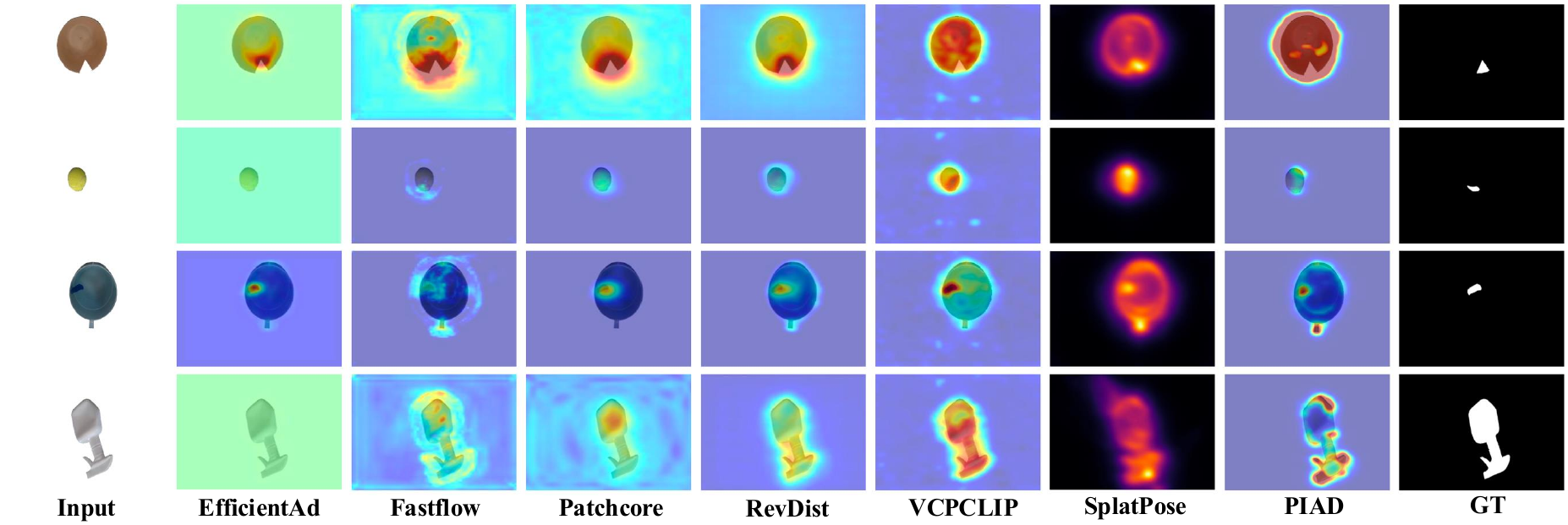}
    \caption{\textbf{Visualization of pixel-wise anomaly detection baselines.} Heatmaps illustrate the ground truth and inference results for two anomalous objects, bowl and spray bottle.}
    \label{fig:result_qualitative}
\end{figure*}

Qualitative results in \cref{fig:result_qualitative} further support these conclusions. Two-dimensional feature-based methods effectively detect localized texture anomalies such as scratches and stains. Three-dimensional methods suppress some false positives under large pose mismatches by enforcing geometric consistency. Vision-language models often miss true anomalies or generate incorrect detections due to background clutter or ambiguous semantics.

\subsection{Insights and Analysis}

\textbf{Barriers to gains from 3D over 2D.} Our experiments reveal a key observation: in realistic, pose-agnostic anomaly detection scenarios, mature 2D feature embedding methods such as PatchCore and EfficientAD substantially outperform 3D reconstruction-based approaches like SplatPose and PIAD. While neural rendering techniques can generate geometrically consistent reference views, they remain vulnerable to sparse training views, specular reflections, and object symmetries. Under these conditions, reconstruction artifacts are often misinterpreted as anomalies during pixel-wise comparison. In contrast, PatchCore implicitly captures multi-view appearance and geometric variability through its memory bank of normal features, exhibiting greater robustness to minor misalignments and rendering noise.

\textbf{The capability gap of vision-language models.} The consistent underperformance of vision-language models highlights the difficulty of applying large-scale zero-shot models to realistic anomaly detection. Despite their high-level reasoning abilities, commercial models such as Qwen2.5-VL and GPT-4o perform worse than specialized anomaly detectors on RAD. This stems from the absence of pixel-level localization objectives during pretraining, high sensitivity to real-world imaging conditions, and difficulty distinguishing genuine defects from viewpoint-induced appearance changes. Bounding-box approximations introduce additional inaccuracies.

\textbf{Impact of material properties and category-specific challenges.} Detailed analysis shows that detection performance is closely tied to object material and geometry. Categories with strong specular reflectivity or geometric symmetry exhibit significant performance degradation across all methods. Reflective surfaces destabilize feature matching and pose estimation, while symmetry introduces ambiguities in pose optimization, leading to overlooked or mislocalized anomalies. These findings underscore the need for robust pose-agnostic detection systems that combine accurate geometry modeling with anomaly scoring mechanisms capable of tolerating moderate misalignment.

\FloatBarrier
\section{Conclusion}

We introduced RAD, a realistic robot-captured multi-view anomaly detection benchmark that emphasizes pose variation, reflective materials, and viewpoint-dependent visibility. Covering 13 object categories and four defect types, RAD enables systematic evaluation of 2D feature-based methods, 3D reconstruction approaches, and vision-language models.

Our experiments show that mature 2D feature-embedding methods remain strong in pose-agnostic settings, while recent 3D and vision-language approaches still struggle, particularly under sparse views, reflections, and symmetry, despite improved pixel-level localization in some cases. These results indicate that neither naive geometry augmentation nor zero-shot vision-language models are sufficient. Progress will require methods that jointly reason over appearance and geometry with uncertainty awareness, explicitly handle reflectance and symmetry, and remain robust to sparse views and minor calibration errors.

\backmatter

\section*{Data Availability}

The RAD dataset is publicly available at \url{https://huggingface.co/datasets/zhouk777/RAD}, and the benchmark code is available at \url{https://github.com/kaichen-z/RAD}. The statistics in this paper were recounted from the released archives at dataset revision \texttt{5eed55ab545461eb3173ab79e1a468cc0d0f8704}. A per-sample manifest, per-category and per-instance counts, and reproducible statistics scripts are provided in the code repository. Both RGB variants include anomaly masks and camera-pose metadata. Raw depth observations are not released because they are noisy in realistic capture conditions.

\section*{Declarations of Conflict of Interest}

The authors declare that they have no conflicts of interest related to this work.

\bibliography{refs}

\end{document}